\pdfoutput=1
\documentclass[11pt,letterpaper]{article}

\usepackage[T1]{fontenc}
\usepackage[utf8]{inputenc}
\usepackage{lmodern}
\usepackage[margin=1in]{geometry}
\usepackage{amsmath,amssymb}
\usepackage{booktabs}
\usepackage{multirow}
\usepackage{array}
\usepackage{microtype}
\usepackage{enumitem}
\usepackage{xcolor}
\usepackage{tikz}
\usetikzlibrary{positioning,arrows.meta,fit}
\usepackage[numbers,sort&compress]{natbib}
\usepackage{hyperref}
\usepackage{url}

\hypersetup{
  colorlinks=true,
  linkcolor=black,
  citecolor=black,
  urlcolor=blue,
  pdftitle={Adaptive Temporal Retrieval for Time-Sensitive Question Answering},
  pdfauthor={Daksh Agarwal and Shashank Kumar and Rishi Kumar}
}

\title{Adaptive Temporal Retrieval\\for Time-Sensitive Question Answering}

\author{
Daksh Agarwal \qquad Shashank Kumar \qquad Rishi Kumar\\[0.3em]
University School of Automation and Robotics (USAR)\\
Guru Gobind Singh Indraprastha University, Delhi, India
}

\date{November 2026}

\begin{document}
\maketitle

\begin{abstract}
Answers to many questions change over time, yet standard retrieval-augmented generation (RAG) ranks passages by semantic similarity alone and makes no explicit decision about whether to retrieve or which dates the evidence should cover.
TempoRAG adds a temporal policy on top of one dense index: a two-head query router (retrieval necessity; current, historical, or timeless), a date window whose width follows the question's time cue, a sufficiency loop, and date-aware sentence selection.
We compare one frozen configuration with no retrieval, standard RAG, a fixed $\pm 2$-year filter, and a TempRALM-style year-decay ranker, with an extractive reader, answer projection, and entity cut shared by all retrieving systems.
The policy ties the fixed filter on 77 templated questions (exact match 0.948 vs.\ 0.935, McNemar $p=1.0$), beats it on 193 questions about six held-out entities (0.762 vs.\ 0.684, $p=6.1\times10^{-5}$), and has a non-significant containment edge on 500 TimeQA questions (0.296 vs.\ 0.270, $p=0.117$), while clearly beating the decay ranker there (0.234, $p=5.4\times10^{-4}$).
Ablations attribute most of the gains to the entity cut and the sentence selector; the sufficiency loop adds latency but no answers.
Finally, a DeBERTa-v3 router trained on 80{,}145 questions with LLM labels, a year rule, and counterfactual minimal pairs removes the rule router's hand-written trivia list: it matches the rule router's answer correctness on all 770 test questions, raises counterfactual-probe accuracy from 12/36 to 29/36 over a regex-label router, and has a confidence ECE of 0.056, but it does not beat the rules.
\end{abstract}

\section{Introduction}
\label{sec:intro}

A fact such as who holds an office is true on an interval.
The passage nearest to a question in embedding space can be a true sentence about another year.
Retrieval-augmented generation grounds a reader or generator in an external corpus~\citep{lewis2020rag,karpukhin2020dpr}, but the usual pipeline makes no separate decision for ``now'' and ``in 2016''.
Adaptive retrieval methods such as Self-RAG, FLARE, and Adaptive-RAG decide \emph{whether} or \emph{how often} to retrieve~\citep{asai2024selfrag,jiang2023flare,jeong2024adaptiverag}, and temporal retrieval methods such as TempRALM re-rank passages by their distance in time from the query~\citep{gade2024tempralm}.
This work sits between the two: it asks \emph{when} a time-sensitive question needs retrieval and \emph{what} dated evidence it should keep.

The closest prior work, and the one this paper builds on, is TempRALM~\citep{gade2024tempralm}, which adds a temporal relevance term to the retrieval score of a retrieval-augmented language model.
TempoRAG keeps that idea as a baseline (B4, an exponential year-decay ranker over a dense and BM25 pool) and adds what it does not have: a query router that decides whether to retrieve and whether the question is about the present, a dated past, or no time at all; a date window whose width depends on the kind of time cue; and an explicit stopping check.

The contributions are as follows.
\begin{enumerate}[leftmargin=1.6em,itemsep=0.2em]
  \item A fully specified temporal retrieval policy (router, cue-dependent window, sufficiency loop, date-aware sentence selection) and its formulas (Section~\ref{sec:method}).
  \item A controlled comparison in which every retrieving system shares the same reader, answer projection, sentence selector, and entity cut, so that the gaps measure retrieval and not answer formatting, with paired McNemar tests, bootstrap intervals, and one-component-at-a-time ablations on three test sets (Sections~\ref{sec:exp}--\ref{sec:results}).
  \item A learned DeBERTa-v3 router that removes the hand-written trivia whitelist from the retrieval decision, with an analysis of why a first version trained on regex labels failed and how relabelling and minimal pairs fixed it (Section~\ref{sec:learnedrouter}).
  \item An explicit account of the negative and null results: the adaptive window does not beat a fixed window on the templated test, the sufficiency loop does not improve answers, and the learned router ties rather than beats the rules (Section~\ref{sec:limits}).
\end{enumerate}

We call the full system TempoRAG; in the tables it is the row labelled Adaptive.
We do not claim a new state of the art.
Despite the RAG framing, every reported system uses a small extractive reader rather than a generative language model, so the experiments test the retrieval policy, not generation quality.
The templated data are generated, and the TimeQA index is built from the pages attached to the sampled questions rather than from open Wikipedia, so none of the scores below are comparable to published TimeQA results.

\section{Related work}
\label{sec:related}

\paragraph{Retrieval-augmented generation and adaptive retrieval.}
Lewis et al.~\citep{lewis2020rag} condition a generator on passages from a dense index, and dense passage retrieval showed that a dual encoder can replace a sparse first stage for open-domain questions~\citep{karpukhin2020dpr}.
Atlas trains the retriever and the reader together~\citep{izacard2023atlas}.
Several systems decide when to retrieve.
Self-RAG trains a generator to emit reflection tokens that trigger retrieval and critique its output~\citep{asai2024selfrag}.
FLARE retrieves actively during generation when the upcoming sentence contains low-confidence tokens~\citep{jiang2023flare}.
Adaptive-RAG trains a small classifier to choose among no retrieval, single-step retrieval, and multi-step retrieval according to question complexity~\citep{jeong2024adaptiverag}.
Our router plays a similar role, but its classes are temporal (current, historical, timeless) rather than complexity levels, and its retrieval decision is evaluated with an extractive reader rather than a large generator.
We do not reimplement Self-RAG, FLARE, or Adaptive-RAG, so the shared word ``adaptive'' should not be read as a comparison with them.

\paragraph{Time in retrieval.}
Li and Croft put document age into a language model so that recency-seeking queries do not depend on a hand-tuned boost~\citep{li2003time}.
Kanhabua and N{\o}rv{\aa}g compared ranking models that add a temporal similarity to text similarity on a news archive with time-scoped judgments~\citep{kanhabua2011ranking}.
Gade and Jetcheva's TempRALM adds a temporal score to a retrieval-augmented language model's retriever so that passages dated near the query time are preferred~\citep{gade2024tempralm}.
Our B4 baseline takes only that decay idea, $\exp(-|\Delta y|/2)$ on a dense and BM25~\citep{robertson2009bm25} candidate pool; it is not the authors' TempRALM system.
The adaptive window in TempoRAG is a heuristic descendant of these models, not a replacement for them, and we do not evaluate on their collections.

\paragraph{Time in question answering.}
TempQuestions evaluates time expressions against a knowledge graph~\citep{jia2018tempquestions}.
TimeQA aligns time-evolving Wikidata facts with Wikipedia pages and reports exact match; a fine-tuned Fusion-in-Decoder reader reaches 60.5 exact match on its easy test and 46.8 on its hard test, against 87.0 for humans on the hard test~\citep{chen2021timeqa}.
SituatedQA shows that one question can have different answers in different temporal and geographic contexts~\citep{zhang2021situatedqa}.
StreamingQA and RealTime QA test adaptation as new text arrives~\citep{liska2022streamingqa,kasai2023realtimeqa}.
Dhingra et al.\ condition a language model on a timestamp so that closed-book recall of a changing fact depends on the year~\citep{dhingra2022talm}.
FreshLLMs introduces FreshQA, a benchmark of fast-changing and false-premise questions, and a search-augmented prompting method~\citep{vu2023freshqa}.
We cite these benchmarks and models to place the work; we did not run them, and our TimeQA subset is not comparable to their reported scores.

\paragraph{Readers and the first-stage stack.}
SQuAD 2.0 requires a span and an abstention decision~\citep{rajpurkar2018squad2}.
Embeddings are BGE-small~\citep{xiao2024cpack}, searched with an exact FAISS inner-product index~\citep{johnson2019faiss}.
A MiniLM cross-encoder trained on MS MARCO reorders the first-stage hits~\citep{wang2020minilm,nguyen2016msmarco,reimers2019sbert}.
The frozen rule router is DistilBERT~\citep{sanh2019distilbert,devlin2019bert}; the learned router is DeBERTa-v3~\citep{he2023debertav3}, with temporal labels from Qwen2.5-7B-Instruct~\citep{qwen2024qwen25}.
Calibration is reported as expected calibration error~\citep{guo2017calibration}.

\section{Problem setup}
\label{sec:setup}

The corpus is $\mathcal{D}=\{(d_i,\tau_i)\}$.
Here $\tau_i$ is the timestamp stored in the index.
It is not the only date in $d_i$.
On the templated corpus, $\tau_i$ is the date of the fact.
On the TimeQA subset, $\tau_i$ is the first date the tagger finds in the chunk, or 1 January 1970 when it finds none.
A question $q$ may also name a target interval.
The system returns a string $\hat{y}$ and a bit saying whether retrieval ran.

Answer strings are normalised as in SQuAD: lower case, punctuation removed, whitespace collapsed.
Exact match is $1$ when the strings are equal.
Token F1 is the harmonic mean of token precision and token recall, taking the best gold string when several are listed.
Containment is
\begin{equation}
\mathrm{Cont}(\hat{y}, y) =
\begin{cases}
1 & \text{if } y \subseteq \hat{y} \text{ or } \hat{y} \subseteq y,\\
0 & \text{otherwise.}
\end{cases}
\end{equation}
The second branch matters.
A short prediction inside a long gold string scores $1$, and so does a 400-character sentence that merely contains the gold span.
On TimeQA the heuristic answers are sentences, so containment mostly asks whether the selected sentence contains the gold text.
It is not an answer-quality metric.
We report exact match beside it for that reason.

If the question is marked unanswerable, or the gold string is empty, exact match, token F1, and containment are replaced by abstention accuracy:
$1$ when the system abstains on an unanswerable question, and $1$ when it does not abstain on an answerable one.
Abstention means an empty string or a short refusal such as ``insufficient evidence''.
On the 500 TimeQA questions this score is $0.820$ for every system, so it does not separate them.

Precision and recall at $k{=}5$ use the gold document ids stored with the question.
Precision divides the number of gold ids in the retrieved list by the length of that list, capped at 5.
On the templated data the generator stores up to three passages of the named entity, preferring those nearest the asked date.
On TimeQA a gold id is a chunk that contains an answer string, and the list is capped at 8.
A precision at 5 of $0.668$ (Table~\ref{tab:syn77}) is therefore an average hit rate inside the top 5, not evidence that each question has three or four relevant documents.
Questions with an empty gold list are omitted from the average.

Temporal accuracy looks only at $\tau$ of the top document.
For a current question, that timestamp must fall within three years of 1 October 2025.
For a historical question, its year must fall within two of a year written in the question or the time hint.
The metric does not read $\hat{y}$.
It is undefined for TimeQA and is left blank there.
For the fixed-window baseline the metric restates the filter.
That baseline keeps documents whose stored year is within $\pm 2$ of the hinted year, and the metric asks whether the top stored year is within $\pm 2$.
If the filter keeps any document, every survivor passes, and so does the top document.
The score can fall below $1$ only when the filter is empty and the fallback document lies outside the band.
It is not independent evidence that the answer is about the right year.

\section{Method}
\label{sec:method}

\begin{figure}[t]
\centering
\begin{tikzpicture}[
  font=\footnotesize,
  box/.style={draw, rounded corners=2pt, align=center, minimum height=2.1em, inner sep=3pt, fill=white},
  dec/.style={box, fill=gray!12},
  arr/.style={-{Stealth[length=4pt]}, thick},
  node distance=0.45cm and 0.32cm]
\node[box] (q) {Question $q$};
\node[dec, right=of q] (r) {Router\\necessity, time class};
\node[box, above right=0.15cm and 0.32cm of r] (skip) {Skip retrieval\\(read with no documents)};
\node[box, below right=0.15cm and 0.32cm of r] (ret) {Dense top-80 $\to$\\cross-encoder top-40};
\node[dec, right=of ret] (pol) {Temporal policy\\current / window / semantic};
\node[dec, right=of pol] (suf) {Sufficiency\\$S\ge 0.55$?};
\node[box, above=0.55cm of suf] (rd) {Date-aware sentence\\selection + reader};
\draw[arr] (q) -- (r);
\draw[arr] (r) |- node[pos=0.75, above, font=\scriptsize] {skip} (skip);
\draw[arr] (r) |- node[pos=0.75, below, font=\scriptsize] {retrieve} (ret);
\draw[arr] (ret) -- (pol);
\draw[arr] (pol) -- (suf);
\draw[arr] (suf) -- node[right, font=\scriptsize] {yes / budget} (rd);
\draw[arr] (suf.south) -- ++(0,-0.35) -| node[pos=0.25, below, font=\scriptsize] {no: widen window, raise $k$ (at most 2 repairs)} (pol.south);
\draw[arr] (skip.east) -| (rd.north);
\end{tikzpicture}
\caption{The TempoRAG inference path. The rule router skips retrieval only for a fixed trivia list (Section~\ref{sec:when}); the learned router of Section~\ref{sec:learnedrouter} replaces that list. B1--B4 share the reader, answer projection, sentence selector, and entity cut, and differ in the retrieval decision, the window, and the loop.}
\label{fig:pipeline}
\end{figure}
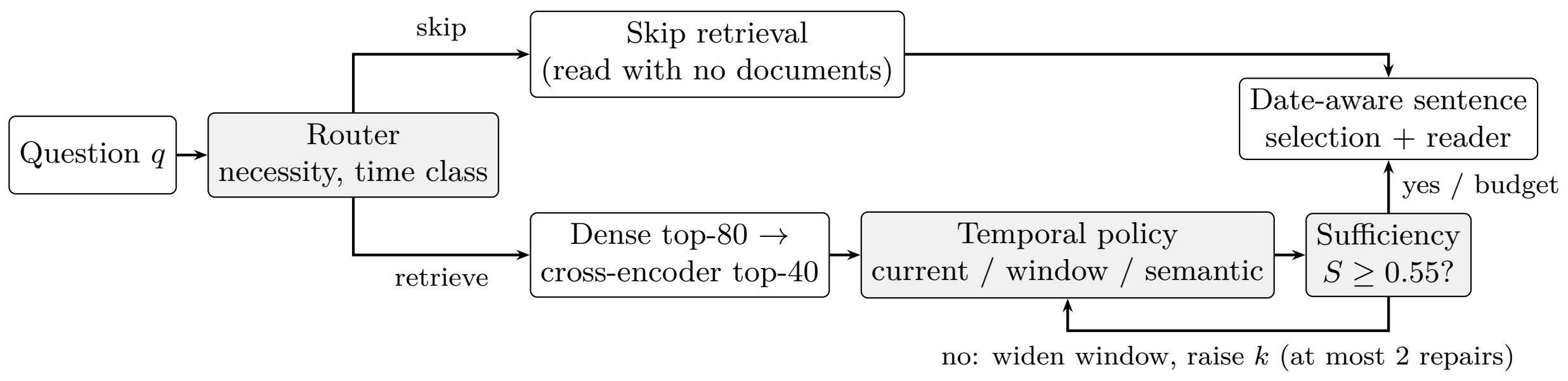

The pipeline classifies the question, retrieves with the dense index, reorders with the cross-encoder, applies a temporal policy, checks sufficiency, and reads an answer.
The index, the cross-encoder, the extractive reader, answer projection, and the entity or surname cut are shared across B1--B3 and Adaptive in the frozen comparison; B4 shares the reader, projection, sentence selector, and entity cut but not the cross-encoder.
The systems differ on whether retrieval can be skipped, how the time window is chosen, and whether a sufficiency loop may retrieve again.
Section~\ref{sec:exp} states that split before any table.
Figure~\ref{fig:pipeline} summarises the flow.

\subsection{Query router}
\label{sec:router}

This subsection describes the frozen rule router used in the main tables.
Section~\ref{sec:learnedrouter} replaces it with a learned router of the same interface.

DistilBERT maps $q$ to the classification-token state $h$.
One linear head emits necessity logits $z^{\mathrm{nec}}\in\mathbb{R}^{2}$.
Another emits temporal logits $z^{\mathrm{tmp}}\in\mathbb{R}^{3}$, over current, historical, and timeless.
A third head emits a confidence logit $z^{\mathrm{conf}}$.
Training uses class weights $w$ and minimises
\begin{equation}
\mathcal{L}
= \mathrm{CE}\!\left(z^{\mathrm{nec}}, y^{\mathrm{nec}}; w^{\mathrm{nec}}\right)
+ \lambda\,
\mathrm{CE}\!\left(z^{\mathrm{tmp}}, y^{\mathrm{tmp}}; w^{\mathrm{tmp}}\right),
\label{eq:loss}
\end{equation}
with $\lambda = 1.75$.
The confidence head is not in $\mathcal{L}$.
Nothing trains its weights, so they stay at initialization.
The encoder still moves, because the two classification losses depend on it.

At inference the class logits are divided by a temperature $T$ before the softmax.
Let $p^{\mathrm{nec}}$ and $p^{\mathrm{tmp}}$ be those distributions, and let $c=\sigma(z^{\mathrm{conf}})$.
The scalar that the policy reads is
\begin{equation}
\gamma =
\begin{cases}
\tfrac12 c + \tfrac12 \max\{\max p^{\mathrm{nec}}, \max p^{\mathrm{tmp}}\}
& \text{when blending is on,}\\[0.3em]
\max\{\max p^{\mathrm{nec}}, \max p^{\mathrm{tmp}}\}
& \text{otherwise.}
\end{cases}
\label{eq:conf}
\end{equation}
The main analyzer blends.
Because $c$ was never trained, half of $\gamma$ is the output of an untrained sigmoid.
If $\gamma < 0.55$, the policy does not trust the temporal class and retrieves in semantic mode.
An untrained head can therefore change which ranker runs.
We did not count how often $\gamma$ falls below $0.55$ on the test sets.
The router of the experimental learned ranking path (Appendix~\ref{sec:learned}) turns blending off, so its $\gamma$ is the maximum class probability.

\subsection{When retrieval is skipped}
\label{sec:when}

Retrieval is skipped only when the necessity class is $0$ and the question string contains one of a fixed list of trivia cues: arithmetic such as ``2 + 2'', ``H2O'', ``days are in a week'', ``spell '', and a few similar items.
A necessity prediction of $0$ on any other question still retrieves, under the name of a soft timeless lookup.
Low confidence also retrieves.
The skip is a whitelist guarded by a classifier, not a learned abstention from retrieval.

There is no parametric model underneath the skip.
The closed-book control is the same extractive reader with an empty document list, plus a few hardcoded replies for those trivia items.
A skip cannot be scored as ``the model already knew the answer.''
It can only be scored as ``the index was not called.''

\subsection{First-stage retrieval}

Documents are embedded with BGE-small English v1.5.
Queries use that model's required prefix.
Search is FAISS \texttt{IndexFlatIP}, exact on these corpora.
Evaluation asks for 80 candidates.
Their FAISS scores are min--max normalised to $\tilde{s}^{\mathrm{f}}_i\in[0,1]$ inside the pool.
The cross-encoder score $s^{\mathrm{ce}}_i$ is left in its raw range.
The blend is
\begin{equation}
s_i = 0.35\,\tilde{s}^{\mathrm{f}}_i + 0.65\,s^{\mathrm{ce}}_i .
\label{eq:blend}
\end{equation}
The top 40 values of $s_i$ are kept.
The two terms do not share a scale, so $0.35$ and $0.65$ are not the weights of a mixture of comparable quantities.
A cross-encoder score of $6$ dominates a dense score in $[0,1]$ whatever the written weights say.
We did not normalise $s^{\mathrm{ce}}$ in the reported runs, and we did not choose the weights on a development split.
They are constants from implementation.

\subsection{Dates}
\label{sec:dates}

Dates are read by rules.
ISO dates and month-year expressions are kept.
A bare year is placed on 15 June, and that midpoint is dropped when the same year already has a fuller date.
Intervals cover $y$--$y$, ``from \ldots to \ldots'', ``until'' or ``through'', two-digit seasons such as 2004--05, and ``for $n$ years beginning in $y$'' with $n\le 40$.
A span longer than 45 years is discarded, which is aimed at birth--death pairs and will also discard a long reign or a long career.
``(born \ldots)'' in the question is stripped before the target year is chosen.
``Before month year'' becomes the interval from three years earlier through that month.
Relative phrases such as ``last year'', ``since 2010'', and ``the predecessor'' are not given their own target.
A decade cue is recognised only when the tagger emits one; there is no general ordinal logic.

Let $I(d)$ be the intervals in a passage and $U(d)$ the point dates, including a usable timestamp.
The distance from a target date $t$ to the passage is
\begin{equation}
\delta(d, t) =
\begin{cases}
0 & \text{if } t \in [u,v] \text{ for some } [u,v]\in I(d),\\[0.25em]
\min\limits_{x \in \partial I(d)\cup U(d)} |x-t| & \text{otherwise,}
\end{cases}
\label{eq:delta}
\end{equation}
where $\partial I(d)$ is the set of endpoints.
Distance zero means the asked date falls inside a stated range.

\subsection{Window and subject filter}

Historical questions use the half-widths in Table~\ref{tab:windows}.
While no surviving document satisfies $\delta(d,t)\le 365.25\,h$, the half-width grows by
\begin{equation}
h \leftarrow \min(h_{\max},\; 1.5h + 0.25).
\label{eq:grow}
\end{equation}
Growth stops at the first in-window survivor or at $h_{\max}$.
We did not ablate the early stop, so we cannot say that stopping early changes the answer.
The widths themselves were set by hand.

\begin{table}[t]
\centering
\caption{Initial and maximum half-width of the historical window, in years.}
\label{tab:windows}
\begin{tabular}{lcc}
\toprule
Cue & Initial $h$ & Maximum $h$ \\
\midrule
Exact calendar date & 0.35 & 1.5 \\
Early, mid, or late in a year & 1.5 & 3.5 \\
A bare year & 1.0 & 4.0 \\
A decade & 5.0 & 8.0 \\
Historical, but no year & 3.0 & 8.0 \\
\bottomrule
\end{tabular}
\end{table}

On the historical branch, before the window test, the pool is cut to a subject.
If any candidate has a non-empty entity field, only documents whose overlap with that field is within $0.01$ of the best overlap are kept.
In the frozen comparison this cut is also applied to B2 and B3, so a precision gap is not an entity-field exclusive.
On the templated corpus the field is the entity name, filled when the passage was generated.
Current mode does not use this cut.
It keeps a lexical tie and sorts by timestamp, and the fixed-window baseline calls that same function when it finds no year.
The overlap text includes the entity string whenever the field is filled, so both systems see it there.
If the field is empty, as it is for every TimeQA chunk, a document is kept when lexical overlap is at least $0.05$ and the text contains a surname, unless overlap is within $0.01$ of the best document.
The surname is the last capitalised token of the first person-like span, ignoring months and question words, and it must be at least three characters.
The rule is English-specific.
It drops a pronoun or a given name, and it was added while inspecting TimeQA failures.

After a stop list, if $H$ remaining query tokens occur in the document, the lexical boost is $b=\min(0.25, 0.05H)$.
The stop list includes months, years, and words such as ``between'', ``held'', and ``team'', because those tokens were matching the wrong passage.

\subsection{Temporal score}

Among documents that survive, semantic scores are min--max normalised to $\tilde{s}_i$.
For current questions, with reference date $r=\text{1 October 2025}$,
\begin{equation}
g^{\mathrm{cur}}_i = \max\left(0,\; 1 - \frac{\max(r-\tau_i,\, 0)}{3650}\right).
\label{eq:cur}
\end{equation}
The denominator is about ten years.
Current mode then keeps documents whose lexical overlap is within $0.01$ of the best and sorts them by $\tau_i$, latest first.
The displayed score of rank $i$ in that tie is $1-0.05\,i/n$, so an older tied document is not treated as weak evidence by the sufficiency check.
``Latest timestamp'' is a proxy for ``current holder'' only when the index date is the fact date.
The reference day is frozen, and it is more than a year before the date of this paper.
No live benchmark is used.

For historical questions,
\begin{equation}
g^{\mathrm{hist}}_i
= \max\left(0,\;
1 - \frac{\delta(d_i, t)}{\max(90,\; 365.25\,w)}
\right),
\label{eq:hist}
\end{equation}
with $w$ the window width in years.
On a tagged corpus, and only when the caller sets a time-first flag, rank is
\begin{equation}
f_i = g^{\mathrm{hist}}_i + 0.01\,\tilde{s}_i + b_i .
\label{eq:timefirst}
\end{equation}
The fixed-window baseline does not set that flag.
On untagged prose with the flag set, a shrunk semantic score is avoided by $\tilde{s}'_i=0.5+0.5\tilde{s}_i$ and
\begin{equation}
f_i = 0.45\,\tilde{s}'_i + 0.55\,g^{\mathrm{hist}}_i + b_i .
\label{eq:untagged}
\end{equation}
Otherwise
\begin{equation}
f_i = \alpha \tilde{s}_i + (1-\alpha)\,g_i\tilde{s}_i + b_i ,
\label{eq:gate}
\end{equation}
with $\alpha=0.6$.
The product $g_i\tilde{s}_i$ is the default, including for the fixed-window baseline's call into the same function.
The top $k=5$ documents go to the reader.
Tagged and untagged questions therefore do not use the same ranking formula.
Saying that one pipeline was applied to every dataset would hide that branch.

\subsection{Sufficiency}

During retrieval the judge sees the min--max semantic score of each document after temporal reordering, not the raw cross-encoder score.
Let $s_{(1)}$ be that score for the top-ranked document, $\overline{s}$ the mean, $C$ the fraction of content query tokens found in the retrieved text, $T\in[0,1]$ a temporal fit, and $b_{(1)}$ the lexical boost of the top document.
\begin{equation}
\begin{aligned}
S =\;& 0.35\,C + 0.25\,\min(1, s_{(1)}) + 0.15\,\overline{s}\\
&+ 0.15\,T + 0.10\,\min(1, b_{(1)}/0.25).
\end{aligned}
\label{eq:suff}
\end{equation}
Because the scores were min--max normalised inside the pool, $s_{(1)}=1$ whenever the temporally top document is also the semantic maximum, which is common.
The $0.25$ term is then a constant.
A pool of poor documents that still differ from each other receives the same saturated value.
$T$ takes the better of the stored timestamp and $\delta(d,t)$, so a passage whose index date is a birth year can still score well if the text contains the asked date.
Evidence is accepted during retrieval when the subject check passes and $S\ge 0.55$.
The weights and the threshold were not fit to data.

If the check fails, the loop may enlarge the candidate pool, raise $k$, add to the window, or fall back toward semantic ranking.
At most two repair rounds run, for three retrieves in total.
A wall-clock budget of 60 seconds per question stops the loop and keeps the best evidence so far.
Both stops are logged.
After an answer is drafted, an abstention stub can trigger one semantic retry only if the budget still allows it, and the retry replaces the answer only when its sufficiency score is strictly higher.

\subsection{Reading the answer}
\label{sec:reader}

The offline reader does not call an external language model.
B1--B4 and Adaptive share this reader and then the same projection.

The shared reader has a few shortcuts.
Trivia cues have fixed replies.
Definition questions return the first sentence.
A passage that contains only one long sentence, and matches an office or winner template, is shortened to the name.
Initials are not sentence breaks, so ``B.\ R.\ Gavai'' stays one token.
That rule was added after a split initial hurt exact match on a templated question.

For a longer passage the reader scores sentences by date agreement, content-word overlap, and a bonus for higher-ranked documents.
A fragment shorter than 24 characters, including a section header, stays attached to the next sentence.
A dated clause shorter than 48 characters is prepended with the previous sentence.
The best sentence is truncated at 400 characters.
This selector is shared.
It was developed against TimeQA failures, including career headers and list sentences.

Projection extracts a name from general who-question and result patterns such as ``held the office'' or ``$X$ won''.
On TimeQA, evaluation also extracts a short span from the selected sentence before exact match and token F1, while containment still uses the sentence.
The patterns do not reliably extract a title such as ``assistant district attorney'' from a sentence that mentions it in passing.

\section{Experiments}
\label{sec:exp}

\subsection{Systems, and what they do not share}

The frozen tables share the extractive reader, answer projection, sentence selector, and entity or surname cut.
B1--B3 and Adaptive differ on the retrieval decision, the time window, and the sufficiency loop.
\begin{itemize}[leftmargin=1.4em]
  \item \textbf{B1.} The extractive reader with no documents. Hardcoded trivia replies still fire. This is not a closed-book language model.
  \item \textbf{B2.} Always retrieve. Rank by \eqref{eq:blend}. No date window and no sufficiency loop.
  \item \textbf{B3.} Always retrieve. If a year is found, keep candidates whose stored year is within $\pm 2$, and if that set is empty keep the unfiltered pool. Then call the shared ranker without the time-first flag. No sufficiency loop.
  \item \textbf{B4.} Dense candidates union BM25 candidates, scored by an equal blend and then multiplied by $\exp(-|\Delta y|/2)$. No hard cutoff and no sufficiency loop. This is a TempRALM-style decay, not the published TempRALM system.
  \item \textbf{Adaptive.} Router, whitelist skip, adaptive window, and sufficiency loop, under the same reader.
  \item \textbf{Adaptive-Fast.} The saved sufficiency-off ablation. Same answers as Adaptive on the templated and held-out files. Labeled as an ablation-derived variant so the negative loop result stays visible. Its latency is from that later run.
\end{itemize}
The experimental learned ranking path of Appendix~\ref{sec:learned} stays in the code, flagged experimental.
It is not a row in the frozen comparison.
The learned router of Section~\ref{sec:learnedrouter} is evaluated by swapping it into this frozen pipeline.

\subsection{Data}

The templated corpus is generated from timelines of office holders, chief executives, sports winners, awards, and a few laws.
Each fact is a short dated passage whose entity field is the timeline name.
The generator writes questions about the holder at a date, the current holder, and timeless trivia, then fills a target size with paraphrases that prepend a stock phrase to an existing question.
The main file has 803 passages and 3{,}000 questions: 2{,}849 train, 74 validation, and 77 test.
Gold document ids are up to three passages of that entity, nearest the asked date.
One test question asks who was president of Brazil in 2016.
The corpus has no passage for the holder in that year.
Every ranker that stays inside 2016 misses it.
That is a hole in the generated data.

The held-out file rebuilds the same generator with a target of 1{,}200 questions, then moves every question that names one of six entities into the test split.
The six names are every fifth entity in a sorted list: Academy Award for Best Picture, CEO of Apple, CEO of Twitter/X, GDPR, President of Brazil, and UEFA Champions League winners.
Test size becomes 193 because those questions are dumped into test, not because a 15\% split was drawn.
Train and validation of this file do not contain the six names.
The analyzer used at test time is the one trained on the main file, and the main file does contain them.
The split therefore does not hold the entities out of the router.
Almost every other component is a rule and was not trained on question labels.
Withholding names from a training split, while evaluating rules, does not test generalisation of a learned retriever.

The TimeQA subset is built from \texttt{hugosousa/TimeQA}.
The loader would take a half-easy, half-hard sample if rows carried a level field.
We did not record whether that field was present, so this paper does not claim an easy or a hard evaluation.
The built subset has 5{,}931 chunks and 900 questions, 400 train and 500 test.
Chunks are pieces of the page attached to the sampled question, not a retrieval corpus over Wikipedia.
A retrieved chunk is a much easier target than a passage from the open encyclopedia.
Gold ids are chunks containing the answer string, at most eight.

\subsection{Frozen configuration}

All numbers in Section~\ref{sec:results} come from one frozen configuration (tagged Checkpoint~B in the repository), \texttt{configs/final.yaml}: extractive reader, $k{=}5$, 80 candidates, 40 kept after the cross-encoder, $\alpha=0.6$, confidence and sufficiency thresholds $0.55$, at most two sufficiency repairs, and a 60-second query timeout.
Sentence selection, date and sentence rules, answer projection, and the entity or surname cut are on.
The experimental learned ranking path is not used by any system in the tables.
\texttt{evaluate.py} loads this file for every system.

Frozen-pipeline runs are on a laptop CPU, with no GPU.
The learned routers of Section~\ref{sec:learnedrouter} were trained on one GPU and evaluated on a different CPU machine, so their latencies are not compared.

The reference date inside \eqref{eq:cur} and inside temporal accuracy is 1 October 2025, the date the templated exam was built against.
Current-question behaviour is measured on that templated exam only.

The main analyzer was trained on the templated training questions with six augmentations, 17{,}094 rows, batch size 8.
The saved log contains 7 epochs.
The configured budget was 15.
We do not have a record of why training stopped, and we did not check whether an augmentation of a validation question occurs in the training rows.
Validation necessity accuracy and temporal accuracy are both $1.0$.
On this data the label is often the word ``current'' or a written year.
The figure says the classifier fits those cues.
It does not say the classifier reads time in Wikipedia prose.
No real-text test of the rule router is reported; Section~\ref{sec:learnedrouter} reports one for the learned routers.

\section{Results}
\label{sec:results}

Tables~\ref{tab:syn77}, \ref{tab:held193}, \ref{tab:timeqa500}, and \ref{tab:ablations} come from the frozen eval directory.
Every system uses the same extractive reader, the same projection, and the same entity or surname cut.

\begin{table}[t]
\centering
\caption{Frozen configuration. Templated test, $n{=}77$. EM is exact match, F1 is token F1, Cont is containment, P@5 and R@5 use gold document ids, Temp is the timestamp metric, Ret is retrieval rate, mean $s$ and med $s$ are seconds.}
\label{tab:syn77}
\small
\begin{tabular}{lccccccccc}
\toprule
System & EM & F1 & Cont & P@5 & R@5 & Temp & Ret & mean $s$ & med $s$ \\
\midrule
B1 no retrieval & 0.130 & 0.167 & 0.130 & 0.000 & 0.000 & 0.247 & 0.000 & 0.000 & 0.000 \\
B2 standard & 0.649 & 0.676 & 0.727 & 0.411 & 0.655 & 0.662 & 1.000 & 1.551 & 1.432 \\
B3 fixed $\pm$2y & 0.935 & 0.951 & 0.987 & 0.608 & 0.943 & 1.000 & 1.000 & 1.555 & 1.452 \\
B4 year decay & 0.662 & 0.684 & 0.727 & 0.435 & 0.695 & 0.675 & 1.000 & 0.031 & 0.030 \\
Adaptive & 0.948 & 0.960 & 0.987 & 0.668 & 0.954 & 1.000 & 0.870 & 1.690 & 1.573 \\
Adaptive-Fast & 0.948 & 0.960 & 0.987 & 0.668 & 0.954 & 1.000 & 0.870 & 1.179 & 1.290 \\
\bottomrule
\end{tabular}
\end{table}

\begin{table}[t]
\centering
\caption{Frozen configuration. Questions naming six entities, $n{=}193$. Columns match Table~\ref{tab:syn77}.}
\label{tab:held193}
\small
\begin{tabular}{lccccccccc}
\toprule
System & EM & F1 & Cont & P@5 & R@5 & Temp & Ret & mean $s$ & med $s$ \\
\midrule
B1 no retrieval & 0.000 & 0.002 & 0.000 & 0.000 & 0.000 & 0.000 & 0.000 & 0.000 & 0.000 \\
B2 standard & 0.332 & 0.361 & 0.409 & 0.346 & 0.577 & 0.596 & 1.000 & 1.569 & 1.489 \\
B3 fixed $\pm$2y & 0.684 & 0.696 & 0.710 & 0.591 & 0.879 & 1.000 & 1.000 & 1.546 & 1.410 \\
B4 year decay & 0.358 & 0.369 & 0.383 & 0.383 & 0.639 & 0.497 & 1.000 & 0.040 & 0.038 \\
Adaptive & 0.762 & 0.766 & 0.762 & 0.682 & 0.912 & 1.000 & 1.000 & 2.584 & 1.715 \\
Adaptive-Fast & 0.762 & 0.766 & 0.762 & 0.682 & 0.912 & 1.000 & 1.000 & 0.663 & 0.649 \\
\bottomrule
\end{tabular}
\end{table}

\begin{table}[t]
\centering
\caption{Frozen configuration. TimeQA test, $n{=}500$. EM and F1 use a span extracted from the selected sentence; Cont uses the sentence. Temp is omitted.}
\label{tab:timeqa500}
\small
\begin{tabular}{lccccccccc}
\toprule
System & EM & F1 & Cont & P@5 & R@5 & Temp & Ret & mean $s$ & med $s$ \\
\midrule
B1 no retrieval & 0.000 & 0.013 & 0.014 & 0.000 & 0.000 & --- & 0.000 & 0.000 & 0.000 \\
B2 standard & 0.006 & 0.077 & 0.260 & 0.235 & 0.468 & --- & 1.000 & 6.056 & 6.118 \\
B3 fixed $\pm$2y & 0.010 & 0.089 & 0.270 & 0.237 & 0.392 & --- & 1.000 & 6.074 & 6.116 \\
B4 year decay & 0.010 & 0.084 & 0.234 & 0.174 & 0.358 & --- & 1.000 & 0.062 & 0.059 \\
Adaptive & 0.008 & 0.088 & 0.296 & 0.241 & 0.488 & --- & 1.000 & 8.736 & 6.492 \\
Adaptive-Fast & 0.008 & 0.088 & 0.298 & 0.243 & 0.491 & --- & 1.000 & 3.184 & 3.160 \\
\bottomrule
\end{tabular}
\end{table}

\begin{table}[t]
\centering
\caption{Ablations of Adaptive on the frozen configuration. Each row turns off one component. Latency is from a later Adaptive-only run and is not matched to Tables~\ref{tab:syn77}--\ref{tab:timeqa500}.}
\label{tab:ablations}
\small
\begin{tabular}{llccccc}
\toprule
Data & Ablation & EM & F1 & Cont & Ret & med $s$ \\
\midrule
\multirow{6}{*}{Templated} & Adaptive (full) & 0.948 & 0.960 & 0.987 & 0.870 & 1.573 \\
 & router off (always retrieve) & 0.948 & 0.960 & 0.987 & 1.000 & 1.456 \\
 & adaptive window off (fixed $\pm$2y) & 0.948 & 0.960 & 0.987 & 0.870 & 1.349 \\
 & sufficiency loop off & 0.948 & 0.960 & 0.987 & 0.870 & 1.290 \\
 & surname/entity cut off & 0.922 & 0.934 & 0.961 & 0.870 & 1.305 \\
 & sentence selector off & 0.948 & 0.960 & 0.987 & 0.870 & 1.276 \\
\midrule
\multirow{6}{*}{Held-out} & Adaptive (full) & 0.762 & 0.766 & 0.762 & 1.000 & 1.715 \\
 & router off (always retrieve) & 0.762 & 0.766 & 0.762 & 1.000 & 0.817 \\
 & adaptive window off (fixed $\pm$2y) & 0.762 & 0.766 & 0.762 & 1.000 & 0.689 \\
 & sufficiency loop off & 0.762 & 0.766 & 0.762 & 1.000 & 0.649 \\
 & surname/entity cut off & 0.699 & 0.703 & 0.699 & 1.000 & 0.670 \\
 & sentence selector off & 0.762 & 0.766 & 0.762 & 1.000 & 0.655 \\
\midrule
\multirow{6}{*}{TimeQA} & Adaptive (full) & 0.008 & 0.088 & 0.296 & 1.000 & 6.492 \\
 & router off (always retrieve) & 0.008 & 0.088 & 0.296 & 1.000 & 3.185 \\
 & adaptive window off (fixed $\pm$2y) & 0.010 & 0.089 & 0.272 & 1.000 & 3.141 \\
 & sufficiency loop off & 0.008 & 0.088 & 0.298 & 1.000 & 3.160 \\
 & surname/entity cut off & 0.006 & 0.076 & 0.252 & 1.000 & 5.430 \\
 & sentence selector off & 0.002 & 0.048 & 0.168 & 1.000 & 4.300 \\
\bottomrule
\end{tabular}
\end{table}

\subsection{Templated test}

Table~\ref{tab:syn77} is the full 77-question templated test.
Adaptive containment is $0.987$, the same as B3: 76 of 77 questions.
Exact match is $0.948$ against $0.935$, and token F1 is $0.960$ against $0.951$.
That is one extra exact match after the reader is shared.
Precision at 5 moves from $0.608$ to $0.668$; recall at 5 is $0.954$ against $0.943$.
The remaining retrieval difference is the adaptive window and the sufficiency loop, not an exclusive entity field.

Retrieval rate $0.870$ is 67 of 77 questions.
The other 10 are the whitelist.
Mean latency is $1.690$\,s for Adaptive and $1.555$\,s for B3; medians are $1.573$\,s and $1.452$\,s.
Skipping those 10 questions does not make Adaptive faster.
B2, which has no date logic, is worse on every answer metric (containment $0.727$, exact match $0.649$).
B4, dense plus BM25 with exponential decay and no hard window, lands next to B2: exact match $0.662$, containment $0.727$.
That gap is the comparison a date filter is supposed to win.
It does not separate the adaptive window from the fixed two-year filter.
Adaptive-Fast, the saved sufficiency-off ablation, matches Adaptive on every answer metric here (exact match $0.948$, containment $0.987$).
Its mean latency in that later run is $1.179$\,s, against $1.690$\,s for the four-system Adaptive row.

Temporal accuracy is $1.000$ for both Adaptive and B3.
For B3 that is the circular reading in Section~\ref{sec:setup}.
Paired McNemar tests and 10{,}000-resample bootstrap intervals are in Section~\ref{sec:sig}.
The exact-match difference versus B3 is one question, and the test does not call it a difference ($p=1.000$).

\subsection{Six entity names}

Table~\ref{tab:held193} is the 193-question file.
Adaptive exact match is $0.762$, against $0.684$ for B3 and $0.332$ for B2.
Containment matches exact match for Adaptive at $0.762$, and is $0.710$ and $0.409$ for B3 and B2.
Token F1 follows: $0.766$, $0.696$, $0.361$.
Precision at 5 is $0.682$ against $0.591$ and $0.346$.
Recall at 5 is $0.912$ against $0.879$ and $0.577$.
Here Adaptive does help relative to a fair fixed window.
The paired exact-match difference versus B3 is $+0.078$ with 95\% interval $[+0.041, +0.119]$ and McNemar $p=6.10\times 10^{-5}$ (15 questions only Adaptive, none only B3).
Containment is $+0.052$ $[+0.021, +0.083]$, $p=0.002$.
B4 is not close: exact match $0.358$, containment $0.383$.
Versus B4, exact match differs by $+0.404$ $[+0.316, +0.492]$, $p=3.06\times 10^{-15}$.
Adaptive-Fast matches the full answer scores (exact match $0.762$) with ablation-run mean latency $0.663$\,s against $2.584$\,s for the four-system Adaptive row.
Retrieval rate is $1$.
The ``when'' decision does nothing on this split.
Mean latency is $2.584$\,s against $1.546$\,s for B3; medians are $1.715$\,s and $1.410$\,s.
The mean--median gap is the sufficiency loop on a minority of questions, not a cost paid on every question.

B3's temporal accuracy is again $1.000$, and so is Adaptive's.
Neither figure is answer quality.

\subsection{TimeQA, 500 questions}

Table~\ref{tab:timeqa500} is the full TimeQA test under the frozen config.
Exact match and token F1 use a span extracted from the selected sentence.
Containment still uses the sentence, because a 400-character sentence that contains the gold string is not an exact match.

Adaptive containment is $0.296$, against $0.270$ for B3 and $0.260$ for B2: 148, 135, and 130 of 500 questions.
Exact match is $0.008$ against $0.010$ and $0.006$: four, five, and three questions.
Token F1 is $0.088$ against $0.089$ and $0.077$.
The extractive span does not turn this reader into a TimeQA-paper number.
What Adaptive changes on this file is recall at 5 ($0.488$ against $0.392$ for B3 and $0.468$ for B2) and a small containment gain.
Precision at 5 is $0.241$, $0.237$, and $0.235$.
Retrieval rate is $1$.
Mean latency is $8.736$\,s against about $6.06$\,s; medians are $6.492$\,s against $6.116$\,s.
B3 remains a weak control on this corpus: stored timestamps are often a birth year or 1 January 1970, so a $\pm 2$ year filter can drop the paragraph that names the asked year.
B4 containment is $0.234$ and exact match is $0.010$, the same exact-match count as B3 (five questions) and one ahead of Adaptive.
The containment difference versus B4 is $+0.062$ $[+0.028, +0.096]$, $p=5.39\times 10^{-4}$.
Versus B3 the containment difference is $+0.026$ $[ -0.004, +0.056 ]$, $p=0.117$, so that gain is not a stable finding.
Exact match versus B3 is $-0.002$, $p=1.000$.
Adaptive-Fast containment is $0.298$ against $0.296$, exact match unchanged at $0.008$, and the ablation-run mean is $3.184$\,s against $8.736$\,s.
The experimental learned ranking path is not in this table.

\subsection{Paired tests}
\label{sec:sig}

\begin{table}[t]
\centering
\caption{Paired tests, Adaptive minus the named system. The interval is a paired bootstrap with 10{,}000 resamples (seed 0); $p$ is an exact two-sided McNemar test on discordant questions.}
\label{tab:sig}
\small
\begin{tabular}{llccr}
\toprule
Data & Comparison & Metric & A$-$B [95\% CI] & $p$ \\
\midrule
Templated & B3 & EM & $+0.013$ $[+0.000,+0.039]$ & 1.000 \\
Templated & B3 & Cont & $+0.000$ $[+0.000,+0.000]$ & 1.000 \\
Templated & B4 & EM & $+0.286$ $[+0.182,+0.390]$ & $2.98\times 10^{-6}$ \\
Held-out & B3 & EM & $+0.078$ $[+0.041,+0.119]$ & $6.10\times 10^{-5}$ \\
Held-out & B3 & Cont & $+0.052$ $[+0.021,+0.083]$ & 0.002 \\
Held-out & B4 & EM & $+0.404$ $[+0.316,+0.492]$ & $3.06\times 10^{-15}$ \\
Held-out & entity cut off & EM & $+0.062$ $[+0.031,+0.098]$ & $4.88\times 10^{-4}$ \\
TimeQA & B3 & Cont & $+0.026$ $[-0.004,+0.056]$ & 0.117 \\
TimeQA & B4 & Cont & $+0.062$ $[+0.028,+0.096]$ & $5.39\times 10^{-4}$ \\
TimeQA & sentence selector off & Cont & $+0.128$ $[+0.084,+0.172]$ & $2.89\times 10^{-8}$ \\
TimeQA & entity cut off & Cont & $+0.044$ $[+0.024,+0.066]$ & $1.13\times 10^{-4}$ \\
TimeQA & window off & Cont & $+0.024$ $[-0.004,+0.052]$ & 0.134 \\
TimeQA & sufficiency off & Cont & $-0.002$ $[-0.006,+0.000]$ & 1.000 \\
\bottomrule
\end{tabular}
\end{table}

Table~\ref{tab:sig} collects the main paired comparisons.

Every interval below is a paired bootstrap of Adaptive minus the named system, 10{,}000 resamples, seed 0.
Every $p$-value is an exact two-sided McNemar test on discordant questions~\citep{mcnemar1947,efron1993bootstrap}.
The full grid, including B2, is \texttt{artifacts/significance.json} in the code release (see Code and data availability).

On the templated test, Adaptive versus B3 containment is a tie ($p=1.000$) and exact match is one question ($p=1.000$).
Versus B4, exact match is $+0.286$ $[+0.182, +0.390]$, $p=2.98\times 10^{-6}$.
On the held-out file the entity-cut ablation is the answer change that survives a paired test: exact match $+0.062$ $[+0.031, +0.098]$, $p=4.88\times 10^{-4}$, 12 questions only the full system.
On TimeQA, sentence-selector off drops containment by $0.128$ $[+0.084, +0.172]$, $p=2.89\times 10^{-8}$.
Entity-cut off drops containment by $0.044$ $[+0.024, +0.066]$, $p=1.13\times 10^{-4}$.
Window off is $+0.024$ $[ -0.004, +0.052 ]$, $p=0.134$.
Sufficiency off is a tie on exact match and one extra contained answer for the ablated system ($0.298$ against $0.296$, $p=1.000$).
The loop is a negative result: it does not buy answers, and the full Adaptive row is the slow one.

\subsection{Ablations}

Table~\ref{tab:ablations} turns off one Adaptive component at a time on each full test: the router (always retrieve), the adaptive window (fixed $\pm 2$ years), the sufficiency loop, the surname or entity cut, and the sentence selector.
The other frozen settings stay on.
Answer scores in this table are comparable.
Latency is from a later Adaptive-only run, so it is not matched to the four-system means in Tables~\ref{tab:syn77}--\ref{tab:timeqa500}.

On the templated and held-out files, turning off the router, the adaptive window, the sufficiency loop, or the sentence selector leaves exact match and containment unchanged.
The entity cut is the component that moves those answers: templated exact match falls from $0.948$ to $0.922$, and held-out exact match from $0.762$ to $0.699$.
Sufficiency off is faster and does not change those scores.

On TimeQA the sentence selector is the largest drop: containment $0.168$ against $0.296$, exact match $0.002$ against $0.008$.
The entity cut also matters (containment $0.252$).
Turning the adaptive window into a fixed $\pm 2$ year filter lowers containment to $0.272$ and recall at 5 to $0.397$, which is the B3 pattern.
Sufficiency off is $0.298$ containment against $0.296$: the loop does not buy answers on this file and it costs time.
The router ablation matches the full system, because retrieval rate is already $1$.

\section{A learned router without the trivia list}
\label{sec:learnedrouter}

The rule router of Section~\ref{sec:router} skips retrieval only when a classifier says ``no'' \emph{and} the question matches a fixed trivia list (Section~\ref{sec:when}).
That list is the weakest part of the ``when to retrieve'' claim.
This section asks whether a learned router can make the same decisions with the list removed.
All learned routers keep the interface of Section~\ref{sec:router} (a necessity head, a three-way temporal head, and a confidence head) and are swapped into the frozen pipeline of Section~\ref{sec:exp} with the sufficiency loop on.
Nothing else in the pipeline changes.

\subsection{A first learned router, and why it failed}

The first learned router is DeBERTa-v3-base~\citep{he2023debertav3} trained on 79{,}982 real questions from a 100{,}000-question pool with regex ``silver'' labels, under which a question needs retrieval whenever it is not timeless.
Its test temporal macro-F1 is $0.997$.
That number measures agreement with the labeller, not temporal understanding: a regex-and-source labeller reproduces all 9{,}941 of its test labels.
On 36 hand-written counterfactual questions (``probe36'': 10 timeless questions that contain a year or a word such as \emph{today}, 10 current questions without a cue word, 10 historical questions without a year, and 6 year-anchored controls) it agrees with the regex labeller on 35 but gets both heads semantically right on only 12.
Swapped into the pipeline without the trivia list, it skips retrieval on 26 of 77 templated questions (the rule router skips 10), including 16 that the rule router answered correctly.
Templated exact match falls to $0.740$, a paired difference of $-0.208$ $[-0.299,-0.117]$, $p=3.05\times10^{-5}$ (16 lost, 0 gained).
Held-out and TimeQA answers are unchanged, because retrieval is never skipped there.

\subsection{Relabelling, a year rule, and minimal pairs}

The retrained routers change the labels and the training data, not the pipeline.
\begin{itemize}[leftmargin=1.4em,itemsep=0.15em]
  \item \textbf{Necessity.} A question has necessity 0 only if it is self-contained (arithmetic, unit conversion, word facts), and 1 for every question drawn from a QA source.
  \item \textbf{Temporal labels.} Qwen2.5-7B-Instruct~\citep{qwen2024qwen25} relabels the temporal class of each real question from its next-token scores for \emph{current}, \emph{historical}, and \emph{timeless}. On real training questions these labels agree with the regex labels 67\% of the time. No human checked them.
  \item \textbf{Minimal pairs.} Generated pairs flip one cue, such as a year or a word like \emph{currently}, and keep the rest of the question.
  \item \textbf{Training.} The (temporal, necessity) cells are sampled with inverse-square-root frequency weights. The confidence head is trained with binary cross-entropy (loss weight 0.3) to predict whether both heads are correct, and class decisions use temperature-scaled probabilities.
\end{itemize}
A second round with only these changes loses one templated answer (exact match $0.935$, $p=1.000$) and two held-out answers ($0.751$, $p=0.500$); TimeQA containment is $0.292$ (5 gained, 7 lost, $p=0.774$).
None of these differences is significant, but all three lost templated and held-out answers ask about a recent past year, such as ``In 2022, who held the position of CEO of Twitter / X?'', and were routed as current.
For each year from 2015 to 2021 the LLM had labelled 17--63\% of the real training questions naming that year as current.

The final round therefore adds a deterministic \emph{year rule}: a real question anchored to an explicit year of 2024 or earlier, where the year is not a quantity or an arithmetic operand, is labelled historical.
The rule changes 3{,}810 training labels (3{,}520 current$\to$historical and 290 timeless$\to$historical), and 274 recent-year minimal pairs are added to the training set.
The final data have 80{,}145 training questions (77{,}157 real and 2{,}988 generated), 10{,}018 validation questions, and 9{,}901 test questions.
Before training, questions that share an exact string, a test entity, or a question template with any of the three test sets or the two probe sets are removed, and a final check finds no overlap.

Two routers are trained on one GPU: DeBERTa-v3-base (3 epochs, learning rate $3\times10^{-5}$, batch size 64, temperature 1.6) and DeBERTa-v3-large (2 epochs, $1\times10^{-5}$, batch size 32, temperature 1.4), both with seed 13 and a maximum length of 64 tokens.
The training loops take 171.4\,s and 406.5\,s.
Each router is trained once; we did not run multiple seeds.

\subsection{End-to-end results}

\begin{table}[t]
\centering
\caption{Learned routers swapped into the frozen pipeline without the trivia list. \emph{Rule} is the Adaptive row of Tables~\ref{tab:syn77}--\ref{tab:timeqa500}. \emph{First} is the regex-label router; \emph{Round 2} adds LLM labels and minimal pairs; \emph{Base} and \emph{Large} are the final round with the year rule. The last column gives the questions gained/lost relative to Rule on the primary metric (EM for templated and held-out, containment for TimeQA) and the exact McNemar $p$; ``0/0'' means no discordant questions. Latency is omitted because the learned-router runs used a different machine.}
\label{tab:learned}
\small
\setlength{\tabcolsep}{4pt}
\begin{tabular}{@{}lcccccc@{}}
\toprule
Router & EM & Cont & P@5 & R@5 & Ret & vs.\ Rule (gained/lost, $p$) \\
\midrule
\multicolumn{7}{@{}l}{\emph{Templated}, $n{=}77$} \\
Rule    & 0.948 & 0.987 & 0.668 & 0.954 & 0.870 & --- \\
First   & 0.740 & 0.779 & 0.578 & 0.833 & 0.662 & 0/16, $3.05\times10^{-5}$ \\
Round 2 & 0.935 & 0.987 & 0.659 & 0.966 & 0.896 & 0/1, 1.000 \\
Base    & 0.948 & 0.987 & 0.668 & 0.954 & 0.909 & 0/0, 1.000 \\
Large   & 0.948 & 0.987 & 0.668 & 0.954 & 0.909 & 0/0, 1.000 \\
\midrule
\multicolumn{7}{@{}l}{\emph{Held-out}, $n{=}193$} \\
Rule    & 0.762 & 0.762 & 0.682 & 0.912 & 1.000 & --- \\
First   & 0.762 & 0.762 & 0.682 & 0.912 & 1.000 & 0/0, 1.000 \\
Round 2 & 0.751 & 0.751 & 0.648 & 0.891 & 1.000 & 0/2, 0.500 \\
Base    & 0.762 & 0.762 & 0.682 & 0.912 & 1.000 & 0/0, 1.000 \\
Large   & 0.762 & 0.762 & 0.680 & 0.914 & 1.000 & 0/0, 1.000 \\
\midrule
\multicolumn{7}{@{}l}{\emph{TimeQA}, $n{=}500$} \\
Rule    & 0.008 & 0.296 & 0.241 & 0.488 & 1.000 & --- \\
First   & 0.008 & 0.296 & 0.242 & 0.489 & 1.000 & 0/0, 1.000 \\
Round 2 & 0.010 & 0.292 & 0.225 & 0.449 & 1.000 & 5/7, 0.774 \\
Base    & 0.008 & 0.296 & 0.239 & 0.483 & 1.000 & 0/0, 1.000 \\
Large   & 0.008 & 0.296 & 0.241 & 0.488 & 1.000 & 0/0, 1.000 \\
\bottomrule
\end{tabular}
\end{table}

Table~\ref{tab:learned} gives the end-to-end comparison.
The final base router gives the same answer correctness as the rule router on every one of the 770 test questions: paired exact-match and containment differences are $+0.000$ with no discordant questions ($p=1.000$) on the templated, held-out, and TimeQA tests, and the large router ties in the same way.
Its paired comparisons with the baselines therefore reproduce those of the rule router: versus B3, held-out exact match is $+0.078$ $[+0.041,+0.119]$, $p=6.10\times10^{-5}$, and versus B4, TimeQA containment is $+0.062$ $[+0.028,+0.096]$, $p=5.39\times10^{-4}$.
The base router skips retrieval on 7 templated arithmetic or unit questions, all of which the rule router also skips, and retrieves for the other 3 trivia-list questions that the rule router skips, still answering them correctly.
Its temporal-class accuracy against the benchmark labels is 1.000, 0.984, and 0.990 on the three tests (large: 1.000, 0.995, and 0.998).
The confidence head's low-confidence fallback ($\gamma<0.55$) never fired on any of the 770 questions for any learned router.

\subsection{Counterfactual probes and calibration}

\begin{table}[t]
\centering
\caption{Counterfactual probes and confidence calibration. \emph{Both}: temporal and necessity heads both correct. Probe36 is the 36-question probe; the challenge set has 72 evaluation-only questions in eight categories. ECE compares the confidence head with both-heads-correct on the router's own test split (15 equal-width bins); for the first router only a validation-split ECE was saved. The year rule also relabels part of the test split, so the Round~2 and final ECEs are not on identical labels.}
\label{tab:probes}
\small
\begin{tabular}{@{}lccccc@{}}
\toprule
Router & Probe36 both & Probe36 temporal & Challenge both & Challenge temporal & ECE \\
\midrule
First   & 12/36 & 12/36 & 27/72 & 44/72 & 0.199 (val) \\
Round 2 & 29/36 & 33/36 & 65/72 & 69/72 & 0.065 \\
Base    & 29/36 & 32/36 & 65/72 & 69/72 & 0.056 \\
Large   & 29/36 & 33/36 & 66/72 & 69/72 & 0.048 \\
\bottomrule
\end{tabular}
\end{table}

Table~\ref{tab:probes} shows where the learned routers differ: not in end-to-end answers, but in what the router has learned.
On probe36 the final routers get both heads right on 29 questions, against 12 for the first router; by category the base router's temporal head is right on 7/10 timeless-with-year questions, 10/10 current-without-cue, 10/10 historical-without-year, and 5/6 controls (first router: 0/10, 3/10, 4/10, and 5/6).
On the 72-question challenge set both heads are right on 65 (base) and 66 (large), against 27 for the first router.
The trained confidence head is reasonably calibrated on the in-distribution test split, with ECE 0.0559 (base) and 0.0478 (large) against 0.199 (validation split) for the first router's head, which was trained against labeller confidence rather than correctness.
Calibration is worse on the small probe set (base: 0.183), so the head should not be trusted out of distribution.
The base router's temporal accuracy on its own test split is 0.906 (macro-F1 0.867; large: 0.912 and 0.874).
Those labels come from the LLM and the year rule, so the figure is not comparable with the first router's 0.997 against regex labels.

The finding is a tie with a cleaner mechanism, not an improvement.
The learned router removes the hand-written list from the retrieval decision without losing an answer and is far more robust to cue-flipping probes, but on these three tests it does not beat the rules.
Section~\ref{sec:limits} lists the ways in which the design of this round was informed by evaluation failures.

\section{Limitations}
\label{sec:limits}

The points below are limits of this study, not future work written as a compliment to the method.

\paragraph{Baselines.}
B4 is a TempRALM-style decay ranker written for this comparison; it is not the authors' TempRALM system~\citep{gade2024tempralm}, and it does not run the cross-encoder, so its latency (0.031\,s, 0.040\,s, and 0.062\,s mean on the three tests) is not matched to B2, B3, or Adaptive.
Self-RAG, FLARE, and Adaptive-RAG~\citep{asai2024selfrag,jiang2023flare,jeong2024adaptiverag} are not reimplemented; a head-to-head comparison with them, ideally with a generative reader, requires new experiments and is the most important missing row.
B1 is the extractive reader with an empty document list plus hardcoded trivia replies, not a closed-book language model.
B3 is a reasonable control when the stored timestamp is the fact date, but on TimeQA 662 of 5{,}931 chunks (11.2\%) carry the 1970-01-01 placeholder and a further 658 (11.1\%) are stamped with the subject's birth year, so a hard window is a weak control there, and B3's perfect temporal accuracy on the templated data largely restates its own filter.

\paragraph{Reader.}
Every system uses a frozen extractive reader with hand-written answer projection, not a generative language model.
The results therefore say how the retrieval policy changes the evidence an extractive reader sees; they do not show that the same policy helps a modern LLM generator, which may also answer some questions from parametric memory.

\paragraph{The ``when'' decision.}
With the rule router, skipping retrieval is a trivia whitelist that fires on 10 of 77 templated questions and on none of the held-out or TimeQA questions, and it does not reduce latency.
The learned router removes the list but skips only 7 templated questions; the held-out and TimeQA tests contain no question on which skipping is correct, so they cannot test the skip decision.
The skip was not tested against a reader that could answer from parametric knowledge.

\paragraph{Design choices made after looking at test failures.}
Constants throughout Section~\ref{sec:method} --- the blend weights, $\lambda$, $\alpha$, the window table, the 45-year cutoff, the 15 June midpoint, the $0.55$ thresholds, and the $0.01$ tie --- were chosen by hand, and some TimeQA rules (the surname cut and parts of the sentence selector) were chosen while inspecting TimeQA failures.
There is no untouched development split for those choices.
For the learned router, the necessity definition was chosen after the 16 templated losses of the first router, and the year rule after the three recent-year losses of the second round.
Both changes are therefore informed by evaluation failures, and the final tie should be read as an optimistic estimate of what this recipe gives on unseen data.

\paragraph{Labels and probes.}
The temporal labels come from Qwen2.5-7B-Instruct without a human check; on real training questions they agree with the regex labels 67\% of the time, and we do not know which source is right on the disagreements.
The challenge set and the minimal-pair augmentation were generated with the authors' own tooling, and the 36-question probe is small and author-labelled; a 29/36 versus 12/36 difference is large, but the per-category counts are single digits.
Each learned router was trained with a single seed.

\paragraph{Data and metrics.}
The templated corpus is generated; one test question asks about a holder in a year for which the corpus has no passage.
The held-out split removes six entity names from that file's training and validation questions, but the rule router used at test time was trained on the main file, which contains them, so the split does not hold the entities out of the router.
The TimeQA index is built from chunks of the pages attached to the 500 sampled questions, not from open Wikipedia, and we did not record whether the sample is balanced between the easy and hard levels.
Containment is bidirectional and, for a sentence-returning reader, close to ``does the selected sentence contain the gold string''; exact match stays at 0.006--0.010 for every retrieving system on TimeQA, so TimeQA does not measure answer extraction here.
The current-question branch is scored against a frozen reference day (1 October 2025) on templated text; FreshQA and RealTime QA were not run.
Reported $p$-values are unadjusted. Applying a Holm--Bonferroni correction to the 13 comparisons in Table~\ref{tab:sig} leaves the same eight comparisons significant at $0.05$.

\section{Conclusion}
\label{sec:conclusion}

TempoRAG makes three retrieval decisions explicit for time-sensitive questions: whether to retrieve, which dates to keep, and when the evidence is sufficient.
With a shared reader, answer projection, and entity cut, the full policy ties a fixed $\pm 2$-year filter on the templated test (one extra exact match, $p=1.0$), beats it on the six-entity held-out test (exact match $0.762$ vs.\ $0.684$, $p=6.1\times10^{-5}$), and has a non-significant containment edge on TimeQA ($0.296$ vs.\ $0.270$, $p=0.117$).
A smooth exponential year decay in the style of TempRALM does not replace either window.
Ablations attribute most of the effect to the entity cut and the date-aware sentence selector; the sufficiency loop, meant to decide when context is enough, adds latency without adding answers.
A DeBERTa-v3 router trained on LLM labels, a year rule, and minimal pairs replaces the hand-written trivia list in the retrieval decision without losing a single answer on 770 test questions, raises counterfactual-probe accuracy from 12/36 to 29/36, and is reasonably calibrated in distribution, but it matches rather than beats the rules.
The main open questions are a head-to-head comparison with Self-RAG and Adaptive-RAG under a generative reader, a test set on which skipping retrieval can actually help, and a human-validated label set.

\section*{Code and data availability}

Code, configurations, the frozen result and significance files, the learned-router training notebook, and the probe and challenge sets are available at
\begin{center}\url{https://github.com/dakshagarwal76/adaptive-temporal-rag-project}\end{center}
Paths below are relative to \path{codebase/adaptive-temporal-rag/} in that repository.
The frozen configuration is \path{configs/final.yaml}, the main tables are \path{artifacts/final_results.md}, and the learned-router numbers are summarised in \path{router_v2/results/published_r4_summary.json}.
Model weights (0.7--1.7\,GB) and the 100{,}000-question training data are not in the repository because of their size.

\bibliographystyle{plainnat}
\bibliography{references}

\appendix

\section{Experimental learned ranking path}
\label{sec:learned}

The code also contains an experimental learned ranking path, a separate stack from the learned router of Section~\ref{sec:learnedrouter}.
It is not the system in any table of this paper and is described only for completeness.

Its router starts from the saved rule-router analyzer, freezes the bottom four of six DistilBERT layers, and trains for one epoch on 240 questions.
Temperature $T=2.8$ was chosen on 48 questions, and the accuracy we report is on those same 48 questions: necessity accuracy $0.812$ (F1 $0.889$) and temporal accuracy $0.812$ (macro-F1 $0.782$).
That is not a held-out estimate.
The drop from the original analyzer's validation accuracy of $1.0$ is also not a like-for-like change.
The original figure is on the templated validation split.
The $0.812$ figure is on this 48-question calibration split.
The original weight file was not overwritten.

The ranker is gradient boosting \cite{friedman2001gbm}, 80 trees of depth 3, on 357 rows of which 143 are positive.
Features include the cross-encoder and FAISS scores, lexical overlap, date distance, the router's class probabilities, and two products: current-probability times recency, and historical-probability times $1/(1+\text{year distance})$.
At inference a temporal prior
\begin{equation}
\pi_i
= p_{\mathrm{cur}}\,\rho_i
+ p_{\mathrm{hist}}\frac{1}{1+\Delta_i}
+ p_{\mathrm{time}}\,\tilde{s}^{\mathrm{ce}}_i
\label{eq:prior}
\end{equation}
is mixed as
\begin{equation}
f_i = \eta_i\big(0.35\,p^{\mathrm{gb}}_i + 0.60\,\pi_i + 0.05\,\tilde{s}^{\mathrm{ce}}_i\big),
\label{eq:learnedrank}
\end{equation}
where $\rho_i$ is a recency rank, $\Delta_i$ is year distance, and the subject gate $\eta_i$ is $1$ or $0.25$.
The prior, not the trees, carries most of the weight.

The reader is \texttt{deepset/tinyroberta-squad2}.
The span score is the product of the start and end softmax probabilities.
The no-answer score is $1-\sigma(m)$ with the margin $m$ clipped to $[-20,20]$.
A negative margin does not by itself reject the span.
On a multi-sentence passage the reader first sees the date-matched sentence, and a span scoring at least $0.02$ is kept; otherwise the top document is read if its span scores at least $0.05$.
A logistic regression on reader score, no-answer score, and cross-encoder score was fit on 16 labelled rows after standardising the features.
A hand override also keeps a span when the reader score is at least $0.08$ and the cross-encoder score is at least $1$.
This path retrieves 40 candidates and keeps 20, against 80 and 40 for the heuristic systems, so its latency is not comparable even before the model difference.
\end{document}